\documentclass{article}

\usepackage[utf8]{inputenc} 
\usepackage[T1]{fontenc}    
\usepackage{hyperref}       
\usepackage{xcolor}
\definecolor{another_blue}{rgb}{0.0, 0.47, 0.75}
\definecolor{navyblue}{rgb}{0.0, 0.0, 0.5}
\definecolor{myblue}{rgb}{0.0,0.08,1.0}
\definecolor{royalblue}{rgb}{0.04,0.33,0.64 }
\hypersetup{
    colorlinks,
    citecolor=royalblue,
    filecolor=royalblue,
    linkcolor=royalblue,
    urlcolor=royalblue
}

\usepackage{url}            
\usepackage{booktabs}       
\usepackage{amsfonts}       
\usepackage{nicefrac}       
\usepackage{microtype}      
\usepackage{lipsum}
\usepackage{graphicx}
\usepackage{todonotes}
\usepackage[round]{natbib}
\usepackage{authblk}
\usepackage{lineno} 
\usepackage[nomarkers,figuresonly]{endfloat}
\usepackage[final]{changes}
\graphicspath{ {./images/} }

\title{\replaced{Accelerating Material Design with the Generative Toolkit for Scientific Discovery}{Generative Toolkit for Scientific Discovery}}

\author[1]{Matteo Manica}
\author[1]{Jannis Born}
\author[1]{Joris Cadow}
\author[1]{Dimitrios Christofidellis}
\author[1]{Ashish Dave}
\author[1]{Dean Clarke}
\author[1]{Yves Gaetan Nana Teukam}
\author[1]{Giorgio Giannone}
\author[1]{Samuel C. Hoffman}
\author[1]{Matthew Buchan}
\author[1]{Vijil Chenthamarakshan}
\author[1]{Timothy Donovan}
\author[1]{Hsiang Han Hsu}
\author[1]{Federico Zipoli}
\author[1]{Oliver Schilter}
\author[1]{Akihiro Kishimoto}
\author[1]{Lisa Hamada}
\author[1]{Inkit Padhi}
\author[1]{Karl Wehden}
\author[1]{Lauren McHugh}
\author[1]{Alexy Khrabrov}
\author[1]{Payel Das}
\author[1]{Seiji Takeda}
\author[1]{John R. Smith}
\affil[1]{IBM Research}

\begin{document}
\maketitle
\begin{abstract}
With the growing availability of data within various scientific domains, generative models hold enormous potential to accelerate scientific discovery.
They harness powerful representations learned from datasets to speed up the formulation of novel hypotheses with the potential to impact material discovery broadly.
We present the Generative Toolkit for Scientific Discovery (GT4SD).
This extensible open-source library enables scientists, developers, and researchers to train and use state-of-the-art generative models to accelerate scientific discovery focused on material design.
\end{abstract}

\paragraph{Introduction.}
\replaced{
The rapid technological progress in the last centuries has been largely fueled by the success of the scientific method.
However, in some of the most important fields, such as material or drug discovery, the productivity has been decreasing dramatically~\citep{smietana2016trends} and by today
}{Humanity's progress has been characterised by a delicate balance between curiosity and creativity.
Science is no exception, with its long evolution through trial and error.
While remarkably successful, the scientific method can be a slow iterative process that can be inadequate when faced with critical and pressing needs, e.g., the need to swiftly develop drugs and antibiotics or design novel materials and processes to mitigate climate change effects.
Indeed,} it can take almost a decade to discover a new material and cost upwards of \$10–\$100 million.
One of the most daunting challenges in materials discovery is hypothesis generation\replaced{. The reservoir of natural products and their derivatives has been largely emptied~\citep{atanasov2021natural} and bottom-up human-driven hypotheses have shown that}{, where} it is extremely challenging to identify and select novel and useful candidates in search spaces that are overwhelming in size, e.g., the chemical space for drug-like molecules is estimated to contain $>10^{33}$ structures~\citep{polishchuk2013estimation}.

To overcome this problem, in recent years, \added{machine learning-based }generative models\added{, e.g., Variational Autoencoders (VAEs;~\citep{kingma2013auto}), Generative Adversarial Networks (GANs;~\citep{NIPS2014_5ca3e9b1})} have emerged as a practical approach to designing and discovering molecules with desired properties\added{ leveraging different representations for molecular structure, e.g., text-based like SMILES~\citep{weininger1988smiles} and SELFIES~\citep{krenn2020self} or graph-based~\citep{king1983chemical}}.
\replaced{Compared to exhaustive or grid searches, generative}{Generative} models more efficiently and effectively navigate and explore vast search spaces learned from data based on user-defined criteria.
\replaced{Leveraging these approaches in }{With }a series of seminal works~\citep{gomez2018automatic,segler2018generating,jin2018junction, you2018graph, prykhodko2019novo}, research has covered a wide variety of applications of generative models, including design, optimization and discovery of: sugar and dye molecules~\citep{takeda2020molecular}, ligands for specific targets~\citep{zhavoronkov2019deep, chenthamarakshan2020cogmol, born2021data, hoffman2022optimizing}, anti-cancer hit-like molecules~\citep{mendez2020novo, born2021paccmannrl}\replaced{,}{ and} antimicrobial peptides~\citep{das2021accelerated}\added{ and semiconductors~\citep{siriwardane2022generative}}.

At the same time, we have witnessed growing community efforts for developing software packages to evaluate and benchmark \replaced{machine learning}{generative} models and their application in material science.
\added{On the property prediction side, models, data-mining toolkits and benchmarking suites for material property prediction such as CGCNN~\citep{PhysRevLett.120.145301}, \texttt{pymatgen}~\citep{ong2013python}, Matminer~\citep{ward2018matminer} or Matbench/AutoMatminer~\citep{dunn2020benchmarking} were released.}
\replaced{On the generative side, i}{I}nitial efforts for generic frameworks implementing popular baselines and metrics such as GuacaMol~\citep{brown2019guacamol} and Moses~\citep{polykovskiy2020molecular} paved the way for domain-specific generative model software that is gaining popularity in the space of drug discovery
such as TDC\added{ (Therapeutics Data Commons;}~\citet{Huang2021tdc}).

More recently novel families of methods have been proposed. Generative Flow Networks (GFN;~\citep{bengio2021flow, bengio2021gflownet, jain2022biological}), a generative model that leverages ideas from reinforcement learning to improve sample diversity, provides a non-iterative sampling mechanism for structured data over graphs. GFNs are particularly suited for molecule generation, where sample diversity is challenging.
Diffusion models (DM;~\citep{sohl2015deep, song2019generative, ho2020denoising}) are generative models that learn complex high-dimensional distributions denoising the data at multiple scales. DMs achieve impressive results in terms of sample quality and diversity for unconditional and conditional vision tasks. Recently, text-conditional diffusion models~\citep{ramesh2022hierarchical, rombach2022high, saharia2022photorealistic} have paved the way for a new age of human-machine interaction. Leveraging such advances in conditioning generative models, DMs have been used in the biological domain for molecule conformation using equivariant graph networks~\citep{hoogeboom2022equivariant}, conditioning on a 2D representation of the molecule to generate the 3D pose in space~\citep{xu2022geodiff}, for protein generation~\citep{anand2022protein, wu2022protein}, and docking~\citep{corso2022diffdock}.

\paragraph{Contribution.}
In this landscape, there is a growing need for libraries and toolkits that can lower the barrier to using generative models.
This need is becoming significantly more pressing given the growing models' size and \replaced{their}{companion} significant requirements on considerable computational resources for training them.
This trend \replaced{creates an imbalance between}{effectively excludes large parts of the scientific community from the ability to achieve significant progress. It concentrates the power on} a small, privileged group of researchers in well-funded institutions\added{ and the rest of the scientific community}, thus impeding open, collaborative, and fair science principles~\citep{probst2022growing}.

\begin{figure}[!htb]
\centering
\includegraphics[width=\linewidth]{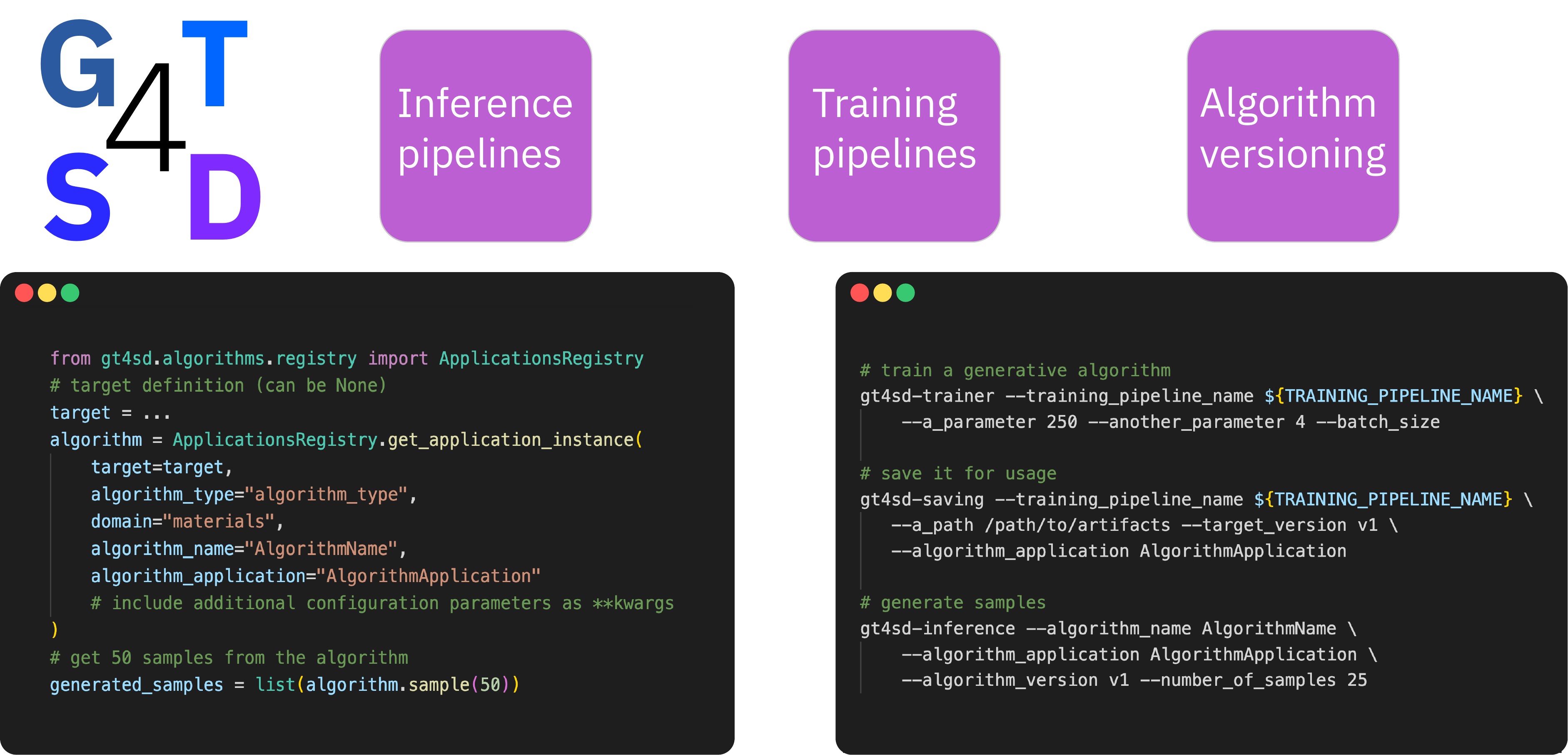}

\vspace{.5cm}

\includegraphics[width=\linewidth]{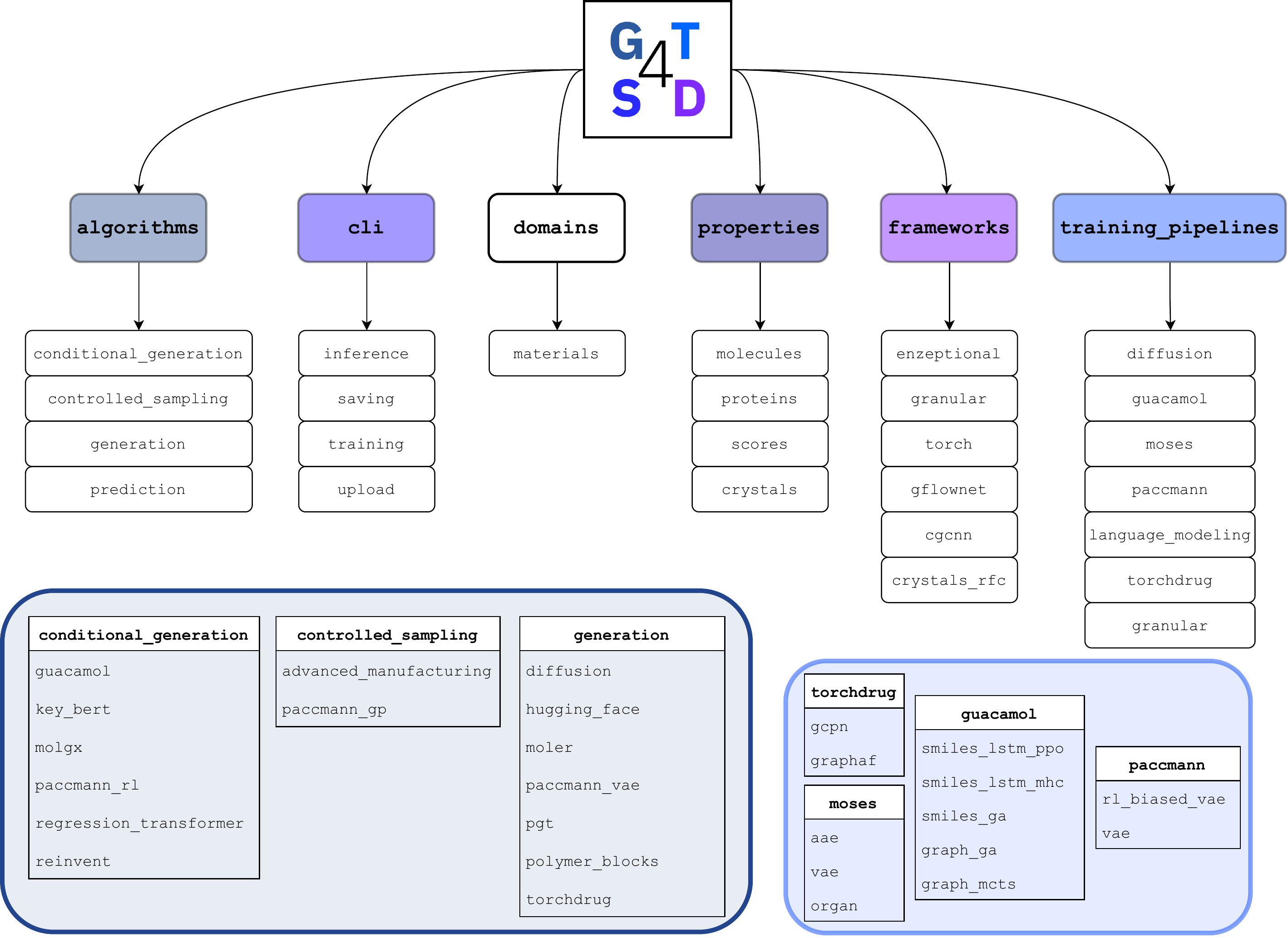}
\caption{
\textbf{GT4SD overview.}
The library implements pipelines for inference and training of generative models. In addition, GT4SD offers utilities for algorithm versioning and sharing for broader usage in the community. The standardised interface enables algorithm instantiation and run for generating samples with less than five lines of code (top, left panel). Furthermore, the CLI tools ease the run of a full discover pipeline in the terminal (top, right panel).
\textbf{GT4SD structure.} 
The library provides (bottom, from left to right) algorithms for inference, a CLI utility, target domains, 
a property prediction interface, 
interfaces and implementations of generative modelling frameworks, 
and training pipelines.
In the blue box, we provide a sample of available frameworks and methodologies for inference algorithms.
}
\label{fig:gt4sd_graphical_abstract}
\end{figure}

We introduce the Generative Toolkit for Scientific Discovery (GT4SD) as a remedy. This python library aims to bridge this gap by developing a framework that eases the training, \replaced{execution}{running}, and develop\replaced{ment}{ing} of generative models to accelerate scientific discovery.
\added{As visualised in~\autoref{fig:gt4sd_graphical_abstract}, GT4SD provides an harmonised interface with a singular application registry for all generative models and a separate registry for properties.
This expenses the need to familiarize with the original developer's code, thus significantly lowering the access barrier. 
Moreover, the high standardization across models eases the integration of new models and facilitates consumption by containerization or distributed computing system.
To the best of our knowledge, GT4SD provides the largest framework for accessing state-of-the-art generative models. It can be used to execute, train, fine-tune and deploy generative models, all either directly through python or via a highly flexible Command Line Interface (CLI).
All pretrained models can be executed directly from the browser through web-apps hosted on Hugging Face Spaces.
Last, for advanced users, the GT4SD model hub simplifies the release of existing algorithms trained on new datasets for instant and continuous integration in their discovery workflows.
}

GT4SD offers a set of capabilities for generating novel hypotheses (inference pipelines) and for fine-tuning domain-specific generative models (training pipelines).
It is designed to be compatible and inter-operable with existing popular libraries, including PyTorch~\citep{NEURIPS2019_9015}, PyTorch Lightning~\citep{Falcon_PyTorch_Lightning_2019}, Hugging Face Transformers~\citep{wolf-etal-2020-transformers}, Diffusers~\citep{von-platen-etal-2022-diffusers}, GuacaMol~\citep{brown2019guacamol}, Moses~\citep{polykovskiy2020molecular}, TorchDrug~\citep{zhu2022torchdrug}, GFlowNets~\citep{jain2022biological} and MoLeR~\citep{maziarz2021learning}.
It includes a wide range of pre-trained models and applications for material design.

GT4SD provides simple interfaces to make generative models easily accessible to users who want to deploy them with just a few lines of code.
The library provides an environment for researchers and students interested in applying state-of-the-art models in their scientific research, allowing them to experiment with a wide variety of pre-trained models spanning a broad spectrum of material science and drug discovery applications.
Furthermore, GT4SD provides a standardised CLI\replaced{,}{ and} APIs for inference and training without compromising on the ability to specify an algorithm's finer-grained parameters\added{ and $>15$ web-apps of various pretrained models}.


\paragraph{Results.}
Arguably, the most considerable potential for accelerating scientific discovery lies in the field of \textit{de novo} molecular design, particularly in material and drug discovery.
With several (pre)clinical trials underway~\citep{jayatunga2022ai}, it is a matter of time until the first AI-generated drug will receive FDA approval and reach the market.
In a seminal study by~\citep{zhavoronkov2019deep}, a deep reinforcement learning model (GENTRL) was utilized for the discovery of potent DDR1 inhibitors, a prominent protein kinase target involved in fibrosis, cancer, and other diseases~\citep{hidalgo2011collective}.
Six molecules were synthesised, four were found active in a biochemical assay, and one compound (in the following called \textit{gentrl-ddr1}) demonstrated favourable pharmacokinetics in mice.
As an exemplary case study in molecular discovery, we consider a contrived task of adapting the hit-compound \textit{gentrl-ddr1} to a similar molecule with an improved \replaced{estimated water solubility (ESOL;~\citet{delaney2004esol}).
Low aqueous solubility affects $>40\%$ of new chemical entities~\citep{savjani2012drug}, thus posing major barriers for drug delivery.
Improving solubility
}{drug-likeness~\mbox{\citep{bickerton2012quantifying}}. 
Quantitative estimate of drug-likeness (or QED) is an \textit{in-silico} property of a molecule that ranges between $0$ and $1$ (\textit{gentrl-ddr1}: $0.38$) and comprises a notion of chemical aesthetics for medicinal chemistry applications.
This task} requires exploring the local chemical space around the hit (i.e., \textit{gentrl-ddr1}) to find an optimized lead compound.

\begin{figure}[!htb]
\centering
\includegraphics[width=\linewidth]{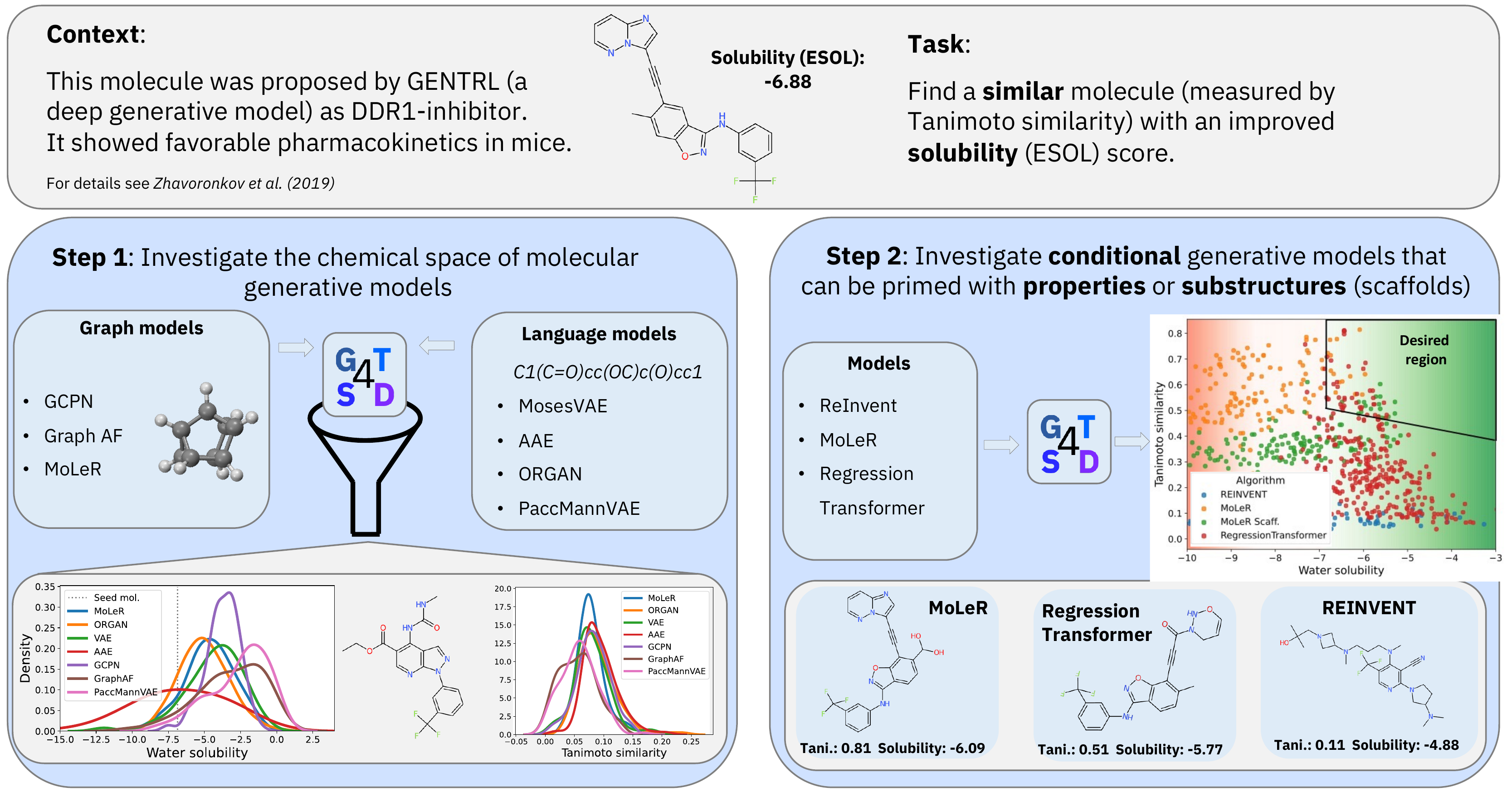}
\caption{
\textbf{Case study using the GT4SD for molecular discovery.}
Starting from a compound designed using generative models by~\citep{zhavoronkov2019deep} (\textit{gentrl-ddr1}), we show how GT4SD can be used to swiftly design molecules with desired properties using a battery of algorithms available in the library in two settings: unconditional (bottom left) and conditional (bottom right).
The conditional models can be  constrained with chemical scaffolds or conditioned on desired property values.
}
\label{fig:casestudy}
\end{figure}

A summary of how this task can be addressed using the GT4SD is shown in~\autoref{fig:casestudy}.
In the first step, a rich set of pre-trained molecular generative models is accessed with the harmonised interface of the GT4SD. 
Two main model classes are available.
The first category is represented by graph generative models, such as MoLeR~\citep{maziarz2022learning} or models from the TorchDrug library,
specifically a graph-convolutional policy network~\citep{you2018graph} and a flow-based autoregressive model (GraphAF;~\citep{shi2020graphaf}).
The second model class is chemical language models (CLM), which treat molecules as text (SMILES~\citep{weininger1988smiles} or SELFIES~\citep{krenn2020self} sequences).
Most of the chemical language models in the GT4SD are accessed via the libraries MOSES~\citep{polykovskiy2020molecular} or GuacaMol~\citep{brown2019guacamol}; in particular a VAE~\citep{gomez2018automatic}, an adversarial autoencoder (AAE;~\citep{kadurin2017cornucopia}) or an objective-reinforced GAN model (ORGAN;~\citep{guimaraes2017objective}).
In the first step, we randomly sample molecules from the learned chemical space of each model.
Assessing the Tanimoto similarity of the generated molecules to \textit{gentrl-ddr1} reveals that this approach while producing many molecules with satisfying \replaced{QED}{ESOL}, did not sufficiently reflect the similarity constraint to the seed molecule (cf.~\autoref{fig:casestudy}, \textit{bottom left}). 
This is expected because the investigated generative models are \textit{unconditional}.

As a more refined approach, the GT4SD includes conditional generative models that can be primed with continuous property constraints or molecular substructures (e.g., scaffolds) such as MoLeR~\citep{maziarz2022learning}, REINVENT~\citep{blaschke2020reinvent} or even with both simultaneously (Regression Transformer;~\citep{born2022regression}).
The molecules obtained from those models, in particular MoLeR and RT, largely respected the similarity constraint and produced many molecules with a Tanimoto similarity $>0.5$ to \textit{gentrl-ddr1}.
\replaced{MoLeR and the RT improved the ESOL by more than $1 M/L$}{The QED constraint was best reflected by the RT, which generated models with an improved QED up to a similarity of $0.82$}
(cf.~\autoref{fig:casestudy}, \textit{right}).
In a realistic discovery scenario, the molecules generated with the described recipes could be manually reviewed by medicinal chemists and selectively considered for synthesis and screening.

\paragraph{GT4SD structure.} 
The GT4SD library follows a modular structure (Figure~\ref{fig:gt4sd_graphical_abstract}) where the main components are: (i) algorithms for serving models in inference mode following a standardised API; (ii) training pipelines sharing a common interface with algorithm families-specific implementations; (iii) domain-specific utilities shared across various algorithms; (iv) a property prediction interface to evaluate generated samples (currently covering small molecules, proteins\added{ and crystals}); (v) frameworks implementing support for complex workflows, e.g., granular for training mixture of generative and predictive models or enzeptional for enzyme design. Besides the core components, there are sub-modules for configuration, handling the cloud object storage-based cache, and error handling at the top-level.

\paragraph{GT4SD inference pipelines.} 
The API implementation underlying the inference pipelines has been designed to support various generative model types: generation, conditional generation, controlled sampling and simple prediction algorithms. All the algorithms implemented in GT4SD follow a standard contract that guarantees a standardised way to call an algorithm in inference mode.
The specific algorithm interface and applications are responsible for defining implementation details and loading the model files from a cache synced with a cloud object storage hosting their versions.

\paragraph{GT4SD training pipelines.} 
Training pipelines follow the same philosophy adopted in implementing the inference pipelines. A common interface allows implementing algorithm family-specific classes with an arbitrary customisable training method that can be configured using a set of data classes. Each training pipeline is associated with a class implementing the actual training process and a triplet of configuration data classes that control arguments for: model hyper-parameters, training parameters, and data parameters.

\paragraph{GT4SD CLI commands.} 
To ease consumption of the pipelines and models implemented in GT4SD, a series of CLI endpoints are available alongside the package: (i) \texttt{gt4sd-inference}, to inspect and run pipelines for inference; (ii) \texttt{gt4sd-trainer}, to list and configure training pipelines; (iii) \texttt{gt4sd-saving}, to persist in a local cache a model version trained via GT4SD for usage in inference mode; (iv) \texttt{gt4sd-upload}, to upload model versions trained via GT4SD on a model hub to share algorithms with other users. 
The CLI commands allow to implement a complete discovery workflow where, starting from a source algorithm version, users can retrain it on custom datasets and make a new algorithm version available in GT4SD.

\paragraph{Discussion.}
The GT4SD is the first step toward a harmonised generative modelling environment for accelerated material discovery.
For the future, we plan to expand application domains (e.g., climate, weather~\citep{ravuri2021skilful}, sustainability, geo-informatics and human mobility~\citep{yan2017universal}), and integrate novel algorithms, ideally with the support of a steadily growing open-science community.

Future developments will focus on two main components: expanding model evaluation and sample properties predictions; developing an ecosystem for sharing models built on top of the functionalities exposed via the existing CLI commands for model lifecycle management.
For the first aspect, we will expand the currently integrated metrics from GuacaMol and Moses and explore bias measures to better analyse performance in light of the generated examples and their properties.
Regarding the sharing ecosystem, we believe GT4SD will further benefit from an intuitive application hub that facilitates distribution of pre-trained generative models (largely inspired by the Hugging Face model hub~\citep{wolf2020transformers}) and enables users to easily fine-tune models on custom data for specific applications.

We anticipate GT4SD to democratise generative modelling in the material sciences and to empower the scientific community to access, evaluate, compare and refine large-scale pre-trained models across a wide range of applications.

\paragraph{Data Availability.}
\added{The complete documentation for the GT4SD code base is available at \url{https://gt4sd.github.io/gt4sd-core/}. Pre-trained models and property predictors are available for automated download via the library itself.}

\paragraph{Code Availability.}
GT4SD source code is available on GitHub: \url{https://github.com/GT4SD/gt4sd-core} (Zenodo DOI: \url{https://zenodo.org/badge/latestdoi/458309249}).
The repository also contains exemplary notebooks and examples for users, including code and data to reproduce the presented case study.
\deleted{The complete documentation is available at \url{https://gt4sd.github.io/gt4sd-core/}}.\added{ Pre-trained generative models and property predictors are also available as Gradio~\citep{Abid_Gradio_Hassle-free_sharing_2019} apps with the corresponding model cards in the GT4SD organization on Hugging Face Spaces: \url{https://huggingface.co/GT4SD}.}

\paragraph{Author Contributions.}
All authors contributed to the design and implementation of different library components before and after its release.
M.M., J.B., D.C., G.G., V.C., A.K., L.M., and J.R.S. contributed to writing and revising the brief communication.
J.B. designed and implemented the case study as well as the Gradio apps.

\paragraph{Competing Interests.}
The authors declare no Competing Financial or Non-Financial Interests.

\paragraph{Acknowledgments.}
\added{
The authors acknowledge Helena Montenegro, Yoel Shoshan, Nicolai Ree and Miruna Cretu for their open-source contributions to the GT4SD.
The authors further thank the anonymous reviewers for their helpful comments.
}
\newpage

\small
\bibliographystyle{plainnat}
\bibliography{references}

\end{document}